\documentclass{article}
\usepackage{ijcai23}

\usepackage{times}
\usepackage{soul}
\usepackage{url}
\usepackage[hidelinks]{hyperref}
\usepackage[utf8]{inputenc}
\usepackage[small]{caption}
\usepackage{graphicx}
\usepackage{amsmath}
\usepackage{amsthm}
\usepackage{booktabs}
\usepackage{bm}
\usepackage{algorithm,algpseudocode}
\usepackage[switch]{lineno}

\usepackage{times}
\usepackage{soul}
\usepackage{url}
\usepackage[hidelinks]{hyperref}
\usepackage[utf8]{inputenc}
\usepackage{graphicx}
\usepackage{amsmath}
\usepackage{amsthm}
\usepackage{booktabs}
\usepackage{csquotes}

\usepackage{amsmath}

\title{Advancing Post-Hoc Case-Based Explanation with Feature Highlighting}

\author{
Eoin M. Kenny$^1$
\and
Eoin Delaney$^{2,4}$\And
Mark T. Keane$^{2,3,4}$
\affiliations
$^1$CSAIL, Massachusetts Institute of Technology\\
$^2$University College Dublin\\
$^3$Insight Centre for Data Analytics\\
$^4$VistaMilk SFI Research Centre
\emails
ekenny@mit.edu,
eoin.delaney@insight-centre.org,
mark.keane@ucd.ie
}
\begin{document}

\maketitle

\begin{abstract}
Explainable AI (XAI) has been proposed as a valuable tool to assist in downstream tasks involving human-AI collaboration. 
Perhaps the most psychologically valid XAI techniques are case-based approaches which display ``whole'' exemplars to explain the predictions of black-box AI systems. 
However, for such \textit{post-hoc} XAI methods dealing with images, there has been no attempt to improve their scope by using multiple clear feature ``parts'' of the images to explain the predictions while linking back to relevant cases in the training data, thus allowing for more comprehensive explanations that are faithful to the underlying model. 
Here, we address this gap by proposing two general algorithms (latent and superpixel-based) which can isolate multiple clear feature parts in a test image, and then connect them to the explanatory cases found in the training data, before testing their effectiveness in a carefully designed user study.
Results demonstrate that the proposed approach appropriately calibrates a user's feelings of ``correctness'' for ambiguous classifications in real world data on the ImageNet dataset, an effect which does not happen when just showing the explanation without feature highlighting.

\end{abstract}

\section{Introduction}
The success of Artificial Neural Networks (ANNs) have led to proposals that they should be used in high-stakes applications such as medical care~\cite{rudin2022interpretable}. 
However, interpretability issues raise questions about their feasibility for such use-cases. 
Accordingly, many eXplainable AI (XAI) techniques have been proposed to overcome this, such as feature highlighting~\cite{ribeiro_why_2016}, and case-based explanations~\cite{papernot2018deep}. 
Case-based Reasoning (CBR) uses training cases directly for inference, thus making it inherently interpretable by way of presenting these cases as explanations~\cite{leake2005introduction}. 
However, methods involving \textit{post-hoc} CBR explanations for image-based ANNs rarely ever consider combining it with feature highlighting, thus allowing explanations to use ``parts'' of images, rather than the whole image, but the ability to do-so would allow explanations to have greater detail, thus allowing more explanatory expression~\cite{chen_this_2018}.
In this paper, to our knowledge, we conduct the first investigation into how to optimally combine CBR explanation with feature-highlighting in a general \textit{post-hoc} manner.
Moreover, we orchestrate the first thorough user evaluation of such explanations.\footnote{Code available at \url{https://github.com/EoinKenny/IJCAI-2023}}

% Furthermore, those few which do consider highlighitng feature parts in images either have poor generalization ability, or do no comparative testing, but simply include it as an ``aside" in the main paper.

% Accordingly, CBR is now widely used to explain ANNs~\cite{kenny2021explaining}. 
% Modernising CBR is one of Rudin's grand challenges for building trustworthy and transparent machine learning systems \cite{rudin2022interpretable}. 
% While there are many CBR approaches to retrieve cases, one of the most glaring issues is a lack of large scale comparative experiments that are supported with user testing. 
% To remedy the situation, this paper conducts a large-scale empirical evaluation to discover the best methods for (1) retrieving explanatory cases, (2) combining them with feature highlighting, and (3) evaluating the effect of such explanations on users.

\begin{figure*}[!t]
  \centering
  \includegraphics[width=\textwidth]{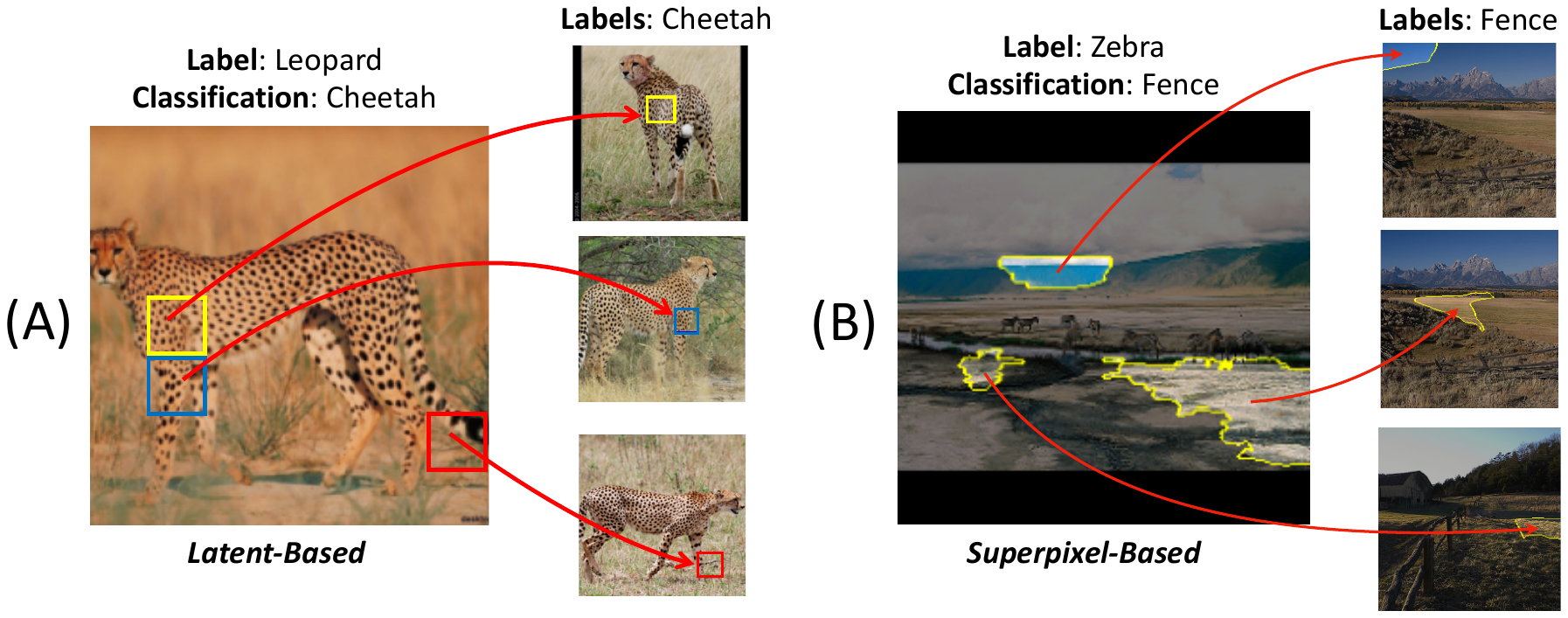} 
  \caption{
  The Algorithms: (A) Our latent-based algorithm explains a misclassification of \enquote{Leopard} as \enquote{Cheetah} from the ImageNet dataset using three feature ``parts'' of the image, and linking them to where they were learned from in the training data.
  The explanation can be parsed as \textit{``I think the image is a Cheetah mainly because of these three features which remind me of Cheetahs that I have seen before''}, thus communicating the classifier got confused because of similar spots and body parts on the two animals.
  (B) The superpixel-based algorithm explains a misclassification of \enquote{Zebra} as \enquote{Fence} on ImageNet in a similar way, where the explanation shows the AI learned a bias correlating \enquote{blue background} and \enquote{grassy fields} with \enquote{Fence}, causing the misclassification, thus communicating the AI has learned bad spurious features to identify the class \enquote{Fence}. 
  }
  \label{fig:title}
\end{figure*}

\section{Related Work}
There are three points we make prior to our literature review which benefit from clarification.
Firstly, XAI can be roughly divided into pre-hoc interpretability and post-hoc explanations~\cite{rudin2022interpretable}, CBR has been used for the former~\cite{kenny2021kbs}, and for the latter~\cite{papernot2018deep}. 
Prior work in pre-hoc interpretability has isolated multiple clear feature ``parts'' in an image by learning prototypical features~\cite{chen_this_2018}, but these techniques often lose accuracy and are highly model specific; we are inspired by these works, but are interested here in developing similar algorithms for \textit{post-hoc} explanation, so that they may be used to explain to any ANN, and never lose model accuracy. 
Secondly, the two prevalent theories in psychology for how humans categorize objects are exemplar and prototype theory~\cite{werner2001categorization}, this is mimicked in the AI literature where CBR  either uses the entire training data for explanations~\cite{kenny2019twin}, or distils it into prototypical examples~\cite{kim2014bayesian}, this paper is only related to the prior literature, and not be be entangled with the latter. 
Thirdly, although our work bears certain resemblance to work in the counterfactual literature~\cite{goyal2019counterfactual}, here we are concerned with similarity-based explanation~\cite{hanawa2021evaluation}, not contrastive.

% \subsection{CBR-XAI Evaluation} 
% The CBR literature either uses the entire training data, or distils it into prototypical examples~\cite{rudin2022interpretable}, this paper is only related to the prior literature, and not be be entangled with the latter. 
% From a computational perspective, Kenny \& Keane~\cite{kenny2021kbs} conducted a large scale evaluation of their case-based explanation technique, but they did not compare to other published methods. 
% Hanawa et al.~\cite{hanawa2021evaluation} also undertook a large-scale empirical investigation into the best similarity metric for retrieving CBR explanations, but they left out several techniques (e.g., see~\cite{jeyakumar2020can}), which we show are important. 
% Hence, there is need for a computational evaluation of all popular published methods to determine the best approach (i.e., Step 1 in Fig.~\ref{fig:title}), which we do in Section~\ref{Section:Expt1}. 

\subsection{CBR and Feature Highlighting}
\label{Section:CBRandCCRsLit}
Work combining feature-highlighting with CBR can be traced back to Patro \& Namboodiri~\shortcite{patro2018differential}, but it was constrained to a specific architecture, and the task of question \& answering, and thus it is not a generalizable solution to the problem we are concerned with. 
In other work, Kenny \& Keane~\shortcite{kenny2019twin} proposed Feature Activation Maps (FAMs), to show similar heatmaps in a test image and nearest neighbor for CBR-based explanation.
However, FAMs are constrained to specific ANN architectures and can not independently isolate \textit{multiple} different features. 
Moreover, FAMs were not comparatively evaluated either computationally (which we rectify in Section~\ref{Section:Expt2}), or in user studies, making it difficult to gauge the utility of FAMs.
Most recently, Crabb\'{e} et al.~\shortcite{crabbe2021explaining} proposed an ANN agnostic method, but it does not highlight regions in a test instance, and it computationally struggles beyond relatively simply domains such as MNIST where explanations are not heavily desired, casting its wider applicability into question. 
Considering this, there is a large gap in the literature for a general, well performing, post-hoc CBR method to attribute multiple feature ``parts'' of an image to corresponding parts of the explanatory cases found.
Moreover, the technique(s) should be thoroughly evaluated in rigorous computational trials to verify the explanation fidelity, which we conduct in Sections~\ref{Section:Expt2} and \ref{Section:Expt3}.

% Considering all this, the only somewhat agnostic technique for our purposes here is FAMs, but notably, the authors have never evaluated the technique in light of recent criticisms of feature attribution techniques \cite{adebayo2018sanity}. 
% Hence, there is need for a computational evaluation of FAMs against other techniques to determine the best approach for the XAI pipeline in question in Figure \ref{fig:title}, which we do in Section~\ref{Section:Expt3}. 
% As an aside, the research here is not to be entangled with Goyal et al.~\shortcite{goyal2019counterfactual}, who produced \textit{counterfactual} explanations, as opposed to \textit{factual} explanations \cite{kenny2021explaining}.

%ED is saying counterfactuals a waste of space. Also do we say not to be confused with too much ... 

\subsection{User Evaluation} 
There is sparse work on user evaluation combining CBR and feature highlighting. 
Rymarczyk et al.~\cite{rymarczyk2021interpretable} continued the work of ProtoPNet~\cite{chen_this_2018} and showed their highlighted features were more distinctive to users. 
However, these authors didn't evaluate the effect of an explanation, but instead whether users found the features distinctive for a class, and crucial details (e.g., number of users) were omitted. 
Crabb\'{e} et al.~\cite{crabbe2021explaining} conducted a user study, but it was really just a pilot (N=10), and there was no significance testing or control.
In light of this, there is a requirement for a well conducted, thorough user trial to test the effectiveness of combining CBR and feature highlighting.
Such a study should assess the effect such explanations have on end-users to ultimately verify their usefulness to people, which we orchestrate in Section~\ref{Section:Expt4}.
% Hence, there is need for an evaluation to see if combining CBR and feature highlighting has a useful effect on humans (i.e., Step 4 in Fig.~\ref{fig:title}), which we do in Section~\ref{Section:Expt4}. 

\subsection{Contributions} 
This paper's main contribution is the proposal of two algorithms in Section~\ref{Section:Method} which can allow the attribution of multiple independent feature ``parts'' of a test image to other parts of the explanatory nearest neighbors found in \textit{post-hoc} CBR exemplar-based explanation (see Figure~\ref{fig:title} for examples of these two algorithms).
As an aside, the evaluation of our algorithms also represents the largest-ever ablation tests comparing popular saliency-based XAI methods in the literature, which used the ILSVRC dataset encompassing 1 Million+ images.
Most importantly however, Section~\ref{Section:Expt4} contributes the first user study purposefully designed to evaluate such explanation strategies in a concrete setting.

\section{Method}
\label{Section:Method}
In this section we introduce our two algorithms for isolating multiple clear feature ``parts'' in a test image, and linking them directly to where they were learned from in the training data.
The first algorithm is an agnostic method for Convolutional Neural Networks (CNNs), and the second is a fully generalizable, agnostic method for all ANNs which uses superpixels.
First however, in preparation, we detail our definitions and assumptions used in the subsequent formalization.

\subsection{Definitions \& Assumptions}
Let a test image be denoted as $I$, and the ANN we wish to explain as $f(.)$.
Assuming a neural net instantiation of this function with a final linear layer, we may decompose $f$ into an encoder $f_{enc}$, alongside the last layer with weights $\bm{W}$ and bias $\bm{b}$ as follows: $f(I) = \bm{W}f_{\text{enc}}(s) + \bm{b}$.
Notably, this encoder $f_{enc}$ outputs a latent vector of the input image $I$, which we denote as $x$, this represents a high level representation of the image.
Hence, we can denote the function as:
\begin{equation}
\begin{gathered}
     f(I) = xW + b.
    \end{gathered}
\end{equation}
\noindent
In the case of CNN architectures, $f_{enc}$ can be further broken down into convolutional layers $g(.)$, and the final transformation operation $t(.)$ which outputs $x$.
The output of $g(.)$ is always a matrix $C\in {\rm I\!R}^{(h,w,d)}$, and the precise operation of $t$ is not important to our formalization, any generic transformation (e.g., a pooling layer or linear transformation) will work.
Thus, in the case of CNNs the function can be written as:
\begin{equation}
\begin{gathered}
     f(I) = t(g(I))W + b \quad \text{where} \quad t(g(I)) = x.
    \end{gathered}
\end{equation}

\noindent
Lastly, we assume the presence of some function $k(.)$ to retrieve a pool of $n$ nearest neighbors of a query $x$ from the training data $D$, all of which are scanned to find where the important feature ``parts'' in the query were learned from which caused the classification.
In theory, the larger $n$ is, the better the explanation will be, as we can have more options to find the best match for the feature(s) in the query image, but due to computational constraints we limited it to $n=50$.

\begin{algorithm}[!b]
\caption{Latent-Based}
\label{alg:cap1}
\begin{algorithmic}[1]
\Require $f(.);$ CNN to-be-explained
\Require $I;$ Test Image
\Require $D;$ Training Dataset
\Require $m(.);$ Activation map algorithm (e.g., FAM)
\State Get Convolutional Output, $C\in {\rm I\!R}^{(h,w,d)} \gets g(I)$                  %\Comment{Get Convolutional Output}
\State Get Activation Map; $M\in {\rm I\!R}^{(h,w)} \gets m(f, I)$                %\Comment{Get Activation Map}
\State Get Pool of $n$ Nearest Neighbors; $\{x_i\}_{i=1}^{n} \gets k(f_{enc}(I), D)$   %\Comment{Get Pool of $k$ Nearest Neighbors}

\State Select the segment $C_{i,j}\in {\rm I\!R}^{(1,1,d)}$ with the maximum saliency $M_{i,j}$ as $\omega_{test}$.

\For{$x_i \in \{x_i\}_{i=1}^{n}$}

    \State $C_x\in {\rm I\!R}^{(h,w,d)} \gets g(x_i)$    
    
    \For{$i$ in range $h$}
    
        \For{$j$ in range $w$}

            \State $\omega_{c} \gets C_{i,j}\in {\rm I\!R}^{(1,1,d)}  $
            \State Record $L_2$ distance $l = \|\omega_{c} - \omega_{test}\|^2_2$
        \EndFor
    \EndFor
\EndFor
\State Select the neighbor $n$ with segment $i, j$ which minimized Eq.~(\ref{eq:latent}).

\end{algorithmic}
\end{algorithm}

\subsection{Latent-Based Algorithm}
\label{section:latentalgo}
The first algorithm takes inspiration from Chen et al.~\shortcite{chen_this_2018}, but in a \textit{post-hoc} (rather than their \textit{pre-hoc}) manner.
%\footnote{Like Patro \& Namboodiri~\cite{patro2018differential}, a comparison to this isn't possible, it requires a specific architecture which cannot explain the pre-trained ResNets used here.} 
% Perhaps most notably, instead of visualizing the explanations as a heatmap, we use \enquote{box} regions like the authors, which allows clear visualization of the features.
For a given test image $I$, a final representation $C\in {\rm I\!R}^{(h,w,d)}$ is extracted after all convolutional layers, where $h$ and $w$ represent the height and width of the output, respectively, and $d$ the number of kernels. 
This may be broken down into regions shaped $h_1 \times w_1 \times d$, where $h_1 < h$ and $w_1 < w$. 
These regions may be upsampled to the size of the test image to visualize them as a \enquote{box} [e.g., Fig.~\ref{fig:expt4}(A)], which is the region in pixel-space that corresponds to this region in $C$.
% \footnote{This could be a different sized box, but like Chen et al.~\cite{chen_this_2018} a small one was chosen to give more granularity.} 
To select salient areas in the test image, the presence of some activation map $M_{test}\in {\rm I\!R}^{(h,w)}$, giving the importance of each spatial region in $C$ is assumed (e.g., FAMs). 
Next, by selecting the most positive salient region(s), one can isolate the \textit{test image} salient box region $\omega_{test} \in {\rm I\!R}^{(1,1,d)}$.

\begin{algorithm}[!b]
\caption{Superpixel-Based}
\label{alg:cap2}
\begin{algorithmic}[1]
\Require $f(.);$ ANN to-be-explained
\Require $I;$ Test Image
\Require $D;$ Training Dataset
\Require $S(.);$ Superpixel Algorithm

\State Get Test Image Superpixels $\{p_i\}_{i=1}^{n} \gets S(I)$
\State Upsample $\{p_i\}_{i=1}^{n}$ 
\State $\{l_i\}_{i=1}^{n} \gets \{  f_{enc}(p_i)  \}_{i=1}^{n}$
\State Get Pool of $n$ Nearest Neighbors; $\{x_i\}_{i=1}^{n} \gets k(f_{enc}(I), D)$

\For{$n_i \in \{x_i\}_{i=1}^{n}$}
    \State $S_n \gets S(x)$
    \For{$s_i \in S_n$}

        \State Occlude $n_i != s_i$
        \State $\hat{y} = f(n_i)$
        \State $\hat{y} = argmax(f_{enc}(I)W + b)$
        \State $m_i = argmax(f_{enc}(s_i)W + b)$
        \State Upsample $s_i$ 
        \State $l_i = f_{enc}(s_i)$
    \EndFor
\EndFor
\State Select neighbor $n$ with segment $l_i$ which minimizes Eq.~(\ref{eq:pixel}).

\end{algorithmic}
\end{algorithm}

Now the task is to find a similar region in the training data $\omega_{nn}$ where the salient feature was \enquote{learned} from. 
In the pool of $n$ nearest neighbors, let $C_{(n, i, j)}$ represent some region $\omega \in {\rm I\!R}^{(1,1,d)}$ in the final convolutional layer, with the neighbor indexed by $n$, and its spatial position in $C_n$ indexed by $i$ and $j$. 
These $n$ images have their $C$ representation searched to find the closest match to $\omega_{test}$ using the $L_2$ norm to find $\omega_{nn}$. 
Crucially, the feature is constrained to its relative importance. 
Specifically, considering each NN's $M_n$, only those regions which satisfy the constraint of being higher than $max(M_n) \times \alpha^{-1}$ are considered to find $\omega_{nn}$ by minimizing:
\begin{equation}
\begin{gathered}
     \underset{n, i, j}{\arg\min} \quad  \|\omega_{test} - C_{(n, i, j)}\|_2   \quad \\ \textrm{s.t.}  \quad  M_{(n, i, j)} > max(M_n) \times \alpha^{-1}.
    \label{eq:latent}
    \end{gathered}
\end{equation}
Alpha ensures $\omega_{nn}$ is critical to the classification, but as it is a hyperparameter we computationally explore its optimal value in Section~\ref{Section:Expt3}.

% For $M$, experiments here use CAM~\cite{zhou2016learning}, FAM~\cite{kenny2019twin}, and Random maps. 
% So three variants are tested, although many other options exist for retrieving $M$~\cite{ramaswamy2020ablation}. 

\subsection{Superpixel-Based Method}
\label{section:supermethod}
This second algorithm uses superpixels to realize a more ANN-agnostic method.
The only assumption is that the ANN extracts a latent representation $\vec{x} \in {\rm I\!R}^{(d)}$ in its penultimate layer before a linear output layer. 
This is important because other similar methods are not ANN agnostic~\cite{kenny2019twin,chen_this_2018}, and would fail to explain state-of-the-art Vision Transformers ~\cite{dosovitskiy2020image}. 
Specifically, LIME~\cite{ribeiro_why_2016} is used to find salient regions of a test image, and it's neighbors are scanned to find the closest match.
However, at scale, LIME was too slow to work on the pool of neighbors, so we developed a new method which we now detail.\footnote{Note that it is not feasible to run LIME on the entire training data a-priori. We examined this possibility but found it would take over a year to process the entire ImageNet dataset using the default hyperparameters.}
% Once a test image CCR is isolated, it is upsampled to the ANN's input-size whilst maintaining its aspect ratio alongside occluding the rest of the image, and its representation is acquired from the ANN's penultimate layer. 
% This process is repeated in the pool of NNs, to find the segment (i.e., NN-CCR) most similar to the test image CCR.

Formally, consider a test image's most salient superpixel region $\omega_{test} \in {\rm I\!R}^{(d)}$, where $d$ is the number of extracted features in the penultimate layer. 
To acquire each regions latent representation, (1) all other regions were occluded, (2) the region is upsampled to the ANN's input size whilst keeping the aspect ratio, and (3) it is passed through the network to record it's representation in layer $x$.
To find a matching region in the training data $\omega_{nn}$ where this feature was \enquote{learned}, let $S_{(n, i)}$ be the representation $\omega \in {\rm I\!R}^{(d)}$ of a superpixel segment $i$ in the NN $n$. 
Moreover, let $M_{(n, i)}$ be each region's saliency, which is acquired by (1) occluding the rest of the image, (2) passing it through the network, and (3) recording the logit value in the originally predicted class by the network.
The $n$ images have their $S$ representations searched to find the closest match to $\omega_{test}$ using the $L_2$ norm to find $\omega_{nn}$. 
Importantly, this region is constrained to its relative importance, only regions with saliency higher than $max(M_n) \times \beta^{-1}$ are considered to find $\omega_{nn}$ by minimizing:
\begin{equation}
\begin{gathered}
     \underset{n, i}{\arg\min} \quad  \|\omega_{test} - S_{(n, i)}\|_2   \quad \\
     \textrm{s.t.}  \quad  M_{(n, i)} > max(M_n) \times \beta^{-1}.
    \label{eq:pixel}
    \end{gathered}
\end{equation}
The beta constraint ensures $\omega_{nn}$ is critical to the classification.
However, again as this is a hyperparameter, its value must be justified though rigorous computational evaluation which we do in Section~\ref{Section:Expt3}.

\begin{figure*}[!t]
  \centering
  \includegraphics[width=\textwidth]{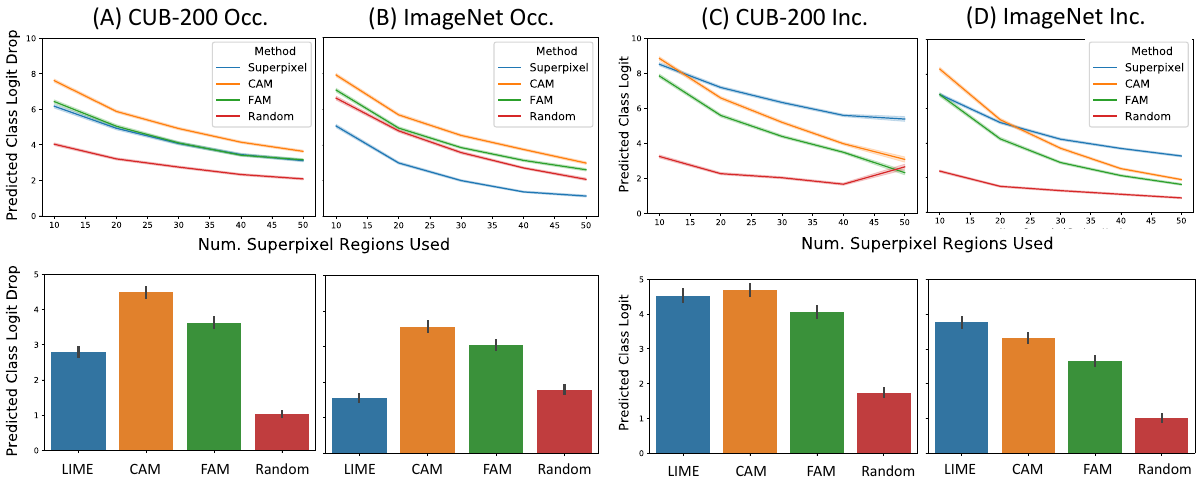} 
  \caption{Expt.~1 Results: Salient Feature Occlusion (Occ.) and Inclusion (Inc.): The first row of lineplots show a comparison of the four different methods proposed in Section~\ref{Section:Method}. The second row shows a comparison of the three latent-based methods against LIME, to see the difference between LIME and our superpixel method. Overall, all methods do significantly better than random occlusion/inclusion, LIME performs similarly to our superpixel method, and CAM performs best. Standard Error bars are shown.}
  \label{fig:expt2}
\end{figure*}

\section{Experiment 1: Test Image Highlighting}
\label{Section:Expt2}
Kenny \& Keane~\shortcite{kenny2019twin} proposed FAMs as (to the best of our knowledge) the only post-hoc method thus far for CBR-based XAI linking an area in the test image to an area in the training one, but the method has two clear drawbacks.
Firstly, FAMs cannot isolate multiple clear feature ``parts'' for an explanation, despite this being argued as a better approach~\cite{chen_this_2018}.
Secondly (and most importantly), FAMs were assumed to be the best method for adding feature highlighting to CBR post-hoc, as no comparative tests were ever performed by the authors, despite recent criticisms of feature attribution methods \cite{adebayo2018sanity,zhou2022feature}.
We address the first issue in our algorithms proposed in Section~\ref{Section:Method} (which allow multiple clear feature parts to be used in an explanation), and the second issue is addressed by comparing FAMs against Class Activation Mapping (CAM), Random maps, LIME~\cite{ribeiro_why_2016}, and our own superpixel method in Section~\ref{section:supermethod} in large a large-scale ablation study next.
% Notably, this is the first large-scale ablation study to compare popular saliency methods such as CAM and LIME in a controlled manner, offering a valuable datapoint for future work.

% proposed FAMs for highlighting one large important region of a test image with a heatmap when combining CBR with feature highlighting.
% However, they did not evaluate its effectiveness at doing so, either on its own or in comparative tests, here we rectify that by comparing FAMs against other popular heatmap methods, namely Class Activation Maps (CAM)~\cite{zhou2016learning}, and random maps.
% Moreover, we perform the first controlled comparisons of such heatmap methods against superpixel methods, namely LIME~\cite{ribeiro_why_2016} and our own superpixel method from Section 3.3 (simply called ``Superpixel" method).
% In doing so, we can isolate the best saliency heatmap/superpixel method to instantiate our algorithms in Section~\ref{Section:Method}.

\subsection{Method and Datasets}
These tests worked via two methods, by (1) keeping the salient region in the pixel image (whilst occluding the rest), and (2) by occluding the salient region (and keeping the rest). 
In the first case, the method that produces the highest logit in the original predicted class does best after a forward pass in the ANN, and in the second the method which produces the highest \textit{drop} is best (n.b., if the prediction changes the same logit is recorded).
For a fair test, the second ablation method was included because the first may be biased towards our Superpixel method which isolate salient regions based on maximizing such logit values [though results show this is not the case, see Fig.~\ref{fig:expt2}(D)]. 
To find equally sized regions between methods, a superpixel segment is first isolated, then the latent-based heatmaps are found by up-sampling their activation maps to the pixel-space, and isolating an equally sized region to the superpixel taken from the parts of highest saliency.
Two datasets CUB-200~\cite{welinder2010caltech} and ImageNet~\cite{5206848} were used, with the former fine-tuning ResNet34, and the latter using a pre-trained ResNet50 (see Appendix).
The experiment is repeated with different segmentation options for superpixels (which changed how big the superpixel regions -- feature parts -- are). 
Tests used the first 500 validation images.

\subsection{Results}
Fig.~\ref{fig:expt2}(A/B) shows the results of occluding (Occ.) the region, and Fig.~\ref{fig:expt2}(C/D) of including (Inc.) it (and occluding the rest of the image). 
The top row shows the results of comparing the four saliency methods discussed in Section~\ref{Section:Method}, whilst the bottom shows comparisons of latent-based methods against LIME. 
Note these LIME comparisons correspond to $\approx$5\% of the test image being isolated, which corresponds to $\approx$30 superpixels used in the top row tests. Overall, all methods are significantly better than random. 
Superpixels perform best for inclusion, especially when the segment number is $>$ 30, but CAMs/FAMs are more consistently good (although FAMs are worse than CAMs). 
Perhaps most notably, all methods perform poorly when occluding in ImageNet, likely because ImageNet has many objects in an image which are used for classification, and removing a small region has little effect. 
All superpixel methods in particular do bad here (i.e., worse than random), likely because they still maintain the \enquote{shape} of the object during occlusion, whilst the latent-based methods (including random) always occlude smoother shapes which distorts the objects more. 
This hypothesis is likely true because this was not repeated in the inclusion experiments, and consistent across all superpixel methods, which is notable because LIME is thought to deliver good explanations~\cite{jeyakumar2020can}. 
So, taking the results as a whole, it is safe to posit that whilst all methods work well in isolating important regions in a test image, CAM does best overall.

\section{Experiment 2: Linking to Training Data}
\label{Section:Expt3}
The purpose of this experiment is to isolate the best hyperparameter values for $\alpha$ and $\beta$ in equations 3 and 4, respectively.
% gather evidence for which highlighted \textit{parts} in the neighboring pool of cases can be attributed to the prediction and hence candidates for explanation (i.e., see Fig.~\ref{fig:title}). 
Knowing these optimal values ensures that the highlighted regions in the explanatory cases are actually relevant to the classification, and not actively misleading end-users.
Hence, potential explanatory features in the nearest neighbors \enquote{used} in classification were occluded in the training data, then the networks were fine-tuned, which allows us to see what regions are necessary to maintain test performance.  
First, $\alpha$ and $\beta$ are varied from one to infinity [the latter considered as all positive salient regions in Eq.~(\ref{eq:latent}/\ref{eq:pixel})] to see their optimal value, then a comparative test is done. 
Eq.~(\ref{eq:latent}) and (\ref{eq:pixel}) cannot be directly compared because alpha and beta are relative constraints. 
However, a comparison was accomplished by (1) varying alpha, (2) closely matching the occlusion area by gradually introducing superpixels (in order of highest saliency value), and (3) readjusting the size of the latent-based methods area to match the superpixel area. 
In essence, we are doing a grid search to find the optimal values for the $\alpha$ and $\beta$ hyperparameters discussed in Section~\ref{Section:Method}, to optimally instantiate our algorithms.

For each hyperparamter value, the networks were fine-tuned for 2500 iterations and test-accuracy sampled every 50, as this was found sufficient for our purposes (20 epochs were tested with no notable differences).
This procedure gives us an indication of \textit{which regions of the training data images} are actually responsible for test predictions (and what $\alpha$ and $\beta$ are best). 
As a sanity check, the networks were also tested by completely occluding all training images and were reduced to random guessing after 2500 iterations, thus verifying that features are being \enquote{unlearned}.
Note this experiment requires a constant value for superpixel segmentation, so 30 was chosen as it was the smallest value which generalized best in Expt.~1.
% Finally, using the best hyperparameters, it was tested how \enquote{similar looking} the NN-CCRs are to the test-image CCR by passing the upsampled regions though the CNN, and comparing their latent representations in the CNN's penultimate layer with the $L_2$ norm. 

\begin{figure*}[!t]
  \centering
  \includegraphics[width=\textwidth]{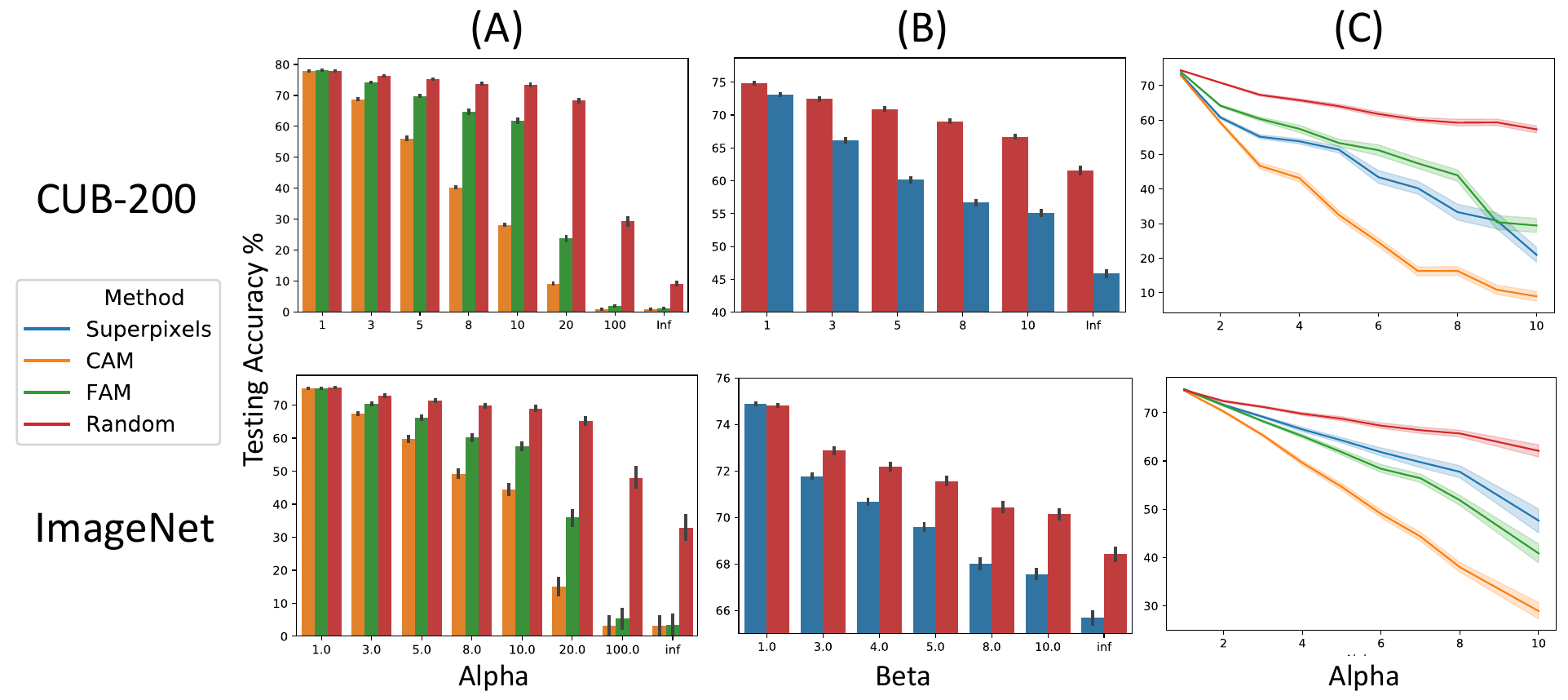} 
  \caption{Expt.~2 Results: (A) Exploring the hyperparameter choice for alpha, (B) doing similar tests for beta. (C) A direct comparison between methods by varying alpha. Results show that all methods are better than the random baseline in nearly every test, FAMs and Superpixels are interchangeable, and CAM seems the best method.}
  
  \label{fig:expt3}
\end{figure*}

\subsection{Results}
Fig.~\ref{fig:expt3}(A) shows $\alpha$ varied, where values between 3-20 produce statistically better results for CAM against FAM/Random (2-tailed ind. t-test; $p$ $<$ 0.05). 
Fig.~\ref{fig:expt3}(B) shows what happens to superpixels v. random when varying $\beta$, where an infinite value (i.e., using all positive superpixels) produces the most divergent results for CUB-200 (Acc. Rand=61.55 v. Superpixel=45.86) and ImageNet (Acc. Rand=68.41 v. Superpixel=65.68). 
Note even with $\beta=inf$ only $\sim$66\% of the images are occluded in ImageNet on average for our superpixel method, which roughly equates to CAM at $\alpha=5$, so there is not a huge disparity when considering this.
Fig.~\ref{fig:expt3}(C) shows a direct comparison, where CAM performs best, with FAMs and superpixels being interchangeable, and all methods outperforming the random baseline. 
% Next, each method's distance of the latent representation between a test image CCR and the NN-CCR was compared using $\beta=\infty$, $\alpha=5$, and the first 500 test images. 
% Results show SP-CCRs are significantly better than other methods when using $>$ 20 segments (mean SP-CCR $L_2$=14.72/13.4 at segments 20/50 v. CAM/FAM mean $L_2$=15.1/15.4 on ImageNet), indicating SP-CCRs find the most \enquote{similar looking} features. 
% Furthermore, the average distance of a neighbor where the latent-based FAM NN-CCR was found was 8.5 cases away (out of $n$=50), this is important because it shows that the method proposed by Kenny \& Keane~\cite{kenny2019twin} will mostly fail to find the best explanation (they only consider $n$=3), whilst our data indicates 95\%+ of the time $n$=17 will find the most similar feature. 
Finally, it should be noted these are empirical tests on retrained networks and are to be treated with some caution, but similar tests have been done and accepted~\cite{hooker2019benchmark}. 
Overall, these results show a converging results that CAM again is the most discriminative.\footnote{Note also that tests were restricted to single ANN architectures because of the enormous computational costs of this experiment. Gathering the data for Fig.~\ref{fig:expt3} took $\approx$ two months on two Nvidia v100 GPUs. Hence, ResNet50 was chosen as it is the most popular for research purposes and helps prove the generality of the results.}

\paragraph{Computational Conclusions.} 
% The results of Expt.~1 showed Twin-Systems the best method for retrieving neighbors. 
Expts.~1-2 showed CAM-based explanations to be best for locating discriminative regions, but superpixels can generalize beyond CNNs~\cite{du2022vision}, and still performed well (but we evaluated them on CNNs so comparative tests were possible). 
Importantly, FAMs~\cite{kenny2019twin} were shown to be less discriminative than CAM, and cannot generalize like Superpixels, thus our tests show a clear improvement upon previously proposed approaches. 
For hyperparameters, latent-based CAM should use $\alpha$=5, as it worked well and has other support~\cite{zhou2016learning}. 
For superpixel methods, LIME should be used to find test image ``parts" (it does slightly better than our superpixel method in Expt.~1), but our superpixel method should be used for finding the matching regions in the training data pool since it performs similarly, is much faster, and has experimental evidence in Expt.~2. 
Superpixel segmentation of 30 and $\beta$=$\infty$ in superpixels is recommended as it generalizes best.
Lastly, note the latent-based algorithm is faster to compute compared to the superpixel one ($\sim$2sec v. $\sim$60; $n$=10; CPU).

\section{User Study}
\label{Section:Expt4}
\begin{figure*}[!t]
  \centering
  \includegraphics[width=\textwidth]{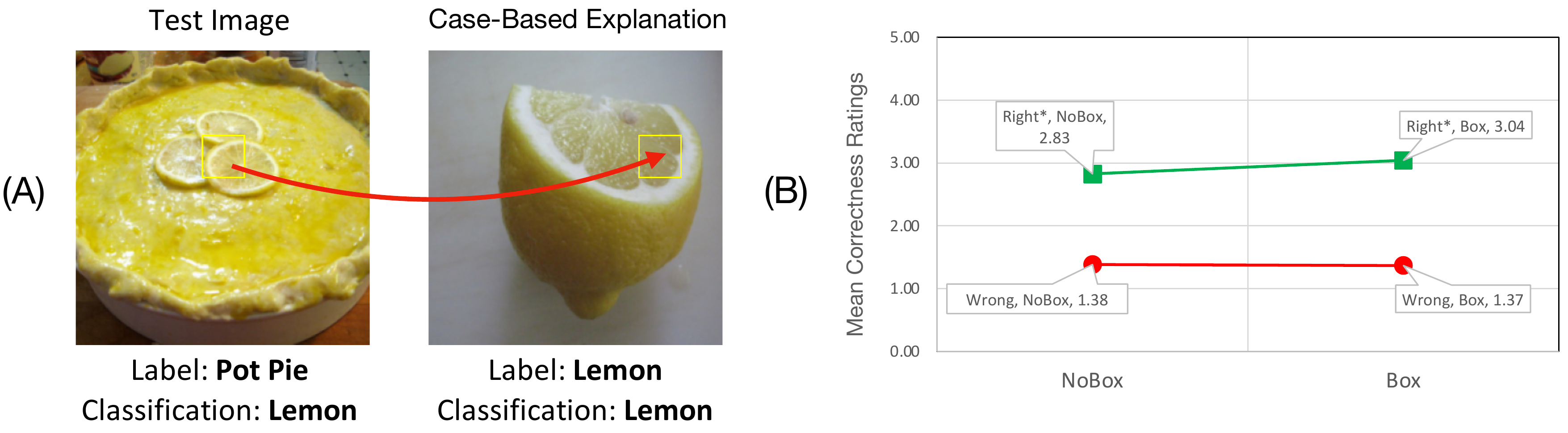} 
  \caption{(A) A \enquote{Pot Pie} misclassified as \enquote{Lemon}. 
  The explanation shows the test image and its relevant salient area alongside its explanatory nearest neighbor in the case-based explanation pointing to where it was \enquote{learned}. 
  Glossed, the explanation says \textit{\enquote{I think this is a lemon, because it has a similar part to an image I saw before which I learned should be a lemon}}. 
  (B) Correctness ratings for material-set B broken out by \enquote{Right*} and Wrong classifications for the two explanation-types, NoBox (example-only) and Box (example-plus-saliency). 
  Here, the Right*-Box ratings (M=3.04) are reliably different to the Right*-NoBox ratings (NoBox, M=2.83), reflecting people's performance on ambiguous items in that material-set.}
  \label{fig:expt4}
\end{figure*}

Although highlighting CBR explanations with important features has become increasingly popular~\cite{chen_this_2018,kenny2019twin,crabbe2021explaining,kenny2021kbs,donnelly2021deformable}, there exists no substantial user study which demonstrates whether such explanations have a useful effect on people.  
So, rather than focusing on comparative testing between various methods, this study examines the more pressing and fundamental question of whether such explanations have a different effect on people than traditional case-based explanation. 
Hence, a user study (N=163) was run to test case-plus-saliency v. case-only explanations, using CAM with $\alpha=5$, a neighbor pool of size $n$=50, and a single highlighted feature. 
Previous similar user studies have shown that case-based explanations changed people's perceptions of the \textit{correctness} of misclassifications~\cite{kenny2021explaining}. 
Hence, here it is examined whether explanatory examples with or without saliency (i.e., a Box v NoBox manipulation; see Fig.~\ref{fig:expt4}) do the same. 
So, the study presented participants with 32 test-images from the ImageNet dataset (i.e., 24 misclassifications, with 8 \enquote{fillers} that were correct classifications for attention checks) and were asked to make classification-correctness judgements of these items presented alongside one of the two explanation-types (NoBox or Box; i.e., no saliency or saliency). 
The 24 misclassifications were randomly divided into two material sets (A-set and B-set) to counterbalance the experiment; so, one group (N=82) received the A-set with case-only explanations (NoBox) and the B-set with case-plus-saliency explanations (Box) and the other group (N=81) received the A-set with case-plus-saliency explanations (Box) and the B-set with case-only explanations (NoBox; see Fig.~\ref{fig:expt4}(A) for example). 
The statistical analysis then collapsed across these counterbalanced groups controlling for the effects of the material-set. 
       
\subsection{Method}
\paragraph{Participants.} Participants (N=163) were recruited on Prolific.co.
All were aged over 18, native English speakers, and lived in the U.S.A., U.K., or Ireland. 
Participants were paid £7.50/hr, which totalled £319.8. 
This N was chosen based on a power analysis for a low effect-size; this size was chosen because it was anticipated the addition of saliency ``boxes" would have a quite nuanced effect over just explanation-by-example due to it already being heavily preferred by users~\cite{jeyakumar2020can}. 
This study passed ethics review of the institution ref. LS-E-19-148-KennyKeane.

\paragraph{Materials.} Twenty-four misclassifications were randomly sampled. 
These were actual test-image errors when the classification differed from the ground truth.  
The twin-system method~\cite{kenny2019twin} was applied to find a nearest-neighbor CBR-explanation, and our latent-based algorithm (i.e., Section~\ref{section:latentalgo}) was used to identify highlighted regions [shown as a Box; see Fig.~\ref{fig:expt4}(A)]. 
The materials were randomly assigned to two different sets (A-set and B-set) and counterbalanced. 
Importantly, the sampling constrained the images to be both varied and those involving classes people could easily understand (e.g., snail, lemon, etc.). 

\paragraph{Procedure.} After being told the system \enquote{learned} to classify objects in images, people were told they would be shown several examples of its classifications (see Appendix). 
Their task was to rate the correctness (i.e., the question was \textit{\enquote{The program's labelling of the image is correct.}}) of the classification on a 5-point Likert-scale from \textit{\enquote{I disagree strongly}} (1) to \textit{\enquote{I agree strongly}} (5). 
Each participant was shown 24 misclassifications (12 NoBox and 12 Box explanations) along with 8 filler items that were all correct classifications appearing every fourth question for attention checks. 
The 24 incorrect items were randomly re-ordered for each person. 
% A debriefing for the study was provided after testing.

%NB removed helpfulness footnote  \footnote{Participants were also asked to rate helpfulness, but the results were not significant and not reported.}

\subsection{Results \& Discussion}
No users failed the attention checks. 
% People perceived the misclassifications as being equally correct for both explanation types; overall, the mean correctness ratings for both explanation-types was the same (M=1.85), and not significantly different (paired t-test; $t$(162) = -0.018, 1-tailed, $p$ $>$ 0.05). 
The analysis showed the items in the B-set that received the example-plus-saliency explanation (Box; M=1.93) were rated as less-incorrect than their equivalent items in Set-A (M=1.77); this difference between B-set-Box and A-set-Box was statistically reliable, $t$(161) = 2.15, $p$ = 0.03, 2-tailed using a two-sample t-test. 
This result shows that image-explanations with feature highlighting impact people's perception of correctness for misclassifications, but only for certain items.\footnote{Interestingly, if counterbalancing was \textit{not} done, one group would show there is \textit{always} a significant increase in correctness for using feature saliency, whilst the other group shows the \textit{complete opposite} in that there is always a significant \textit{decrease} in correctness (due to set B-set being higher in correctness in general). This highlights the need for a controlled counterbalancing design.}

An \textit{ad-hoc} analysis of this effect discovered that in both material-sets there were ambiguous materials that people consistently rated as more correct (i.e., mean $>$ 2) even though the ground-truth identified these items as incorrect (3/12 in A-set and 4/12 in B-set). 
So, the items were partitioned into two new categories, namely \enquote{Right*}-items (that people rated as more \enquote{correct}, even though they were incorrect classifications) and Wrong-items (that people confidently rated as incorrect, when they were incorrect) and then re-analyzed for each material-set (n.b., the asterisk on \enquote{Right*} signifies they are not really \enquote{Right}). 
This partitioning was objectively verified by clustering the material means (using $k$-means) 500 times and finding that the data consistently forms these two groups. 
Fig.~\ref{fig:expt4}(B) shows that in the B-set the correctness rating for the Right*-misclassifications with example-plus-saliency explanation (Box, M=3.04) is reliably higher than that for the example-only explanation (NoBox, M=2.83), $t$(162) = 1.8, $p$ = 0.036, using 1-tailed, two-sample t-test. 
% Note that while this difference appears small on a Likert scale, overall it constitutes a 5.25\% difference in responses, which is a larger effect than that found in other studies~\cite{goyal2019counterfactual}. 
% This shows that some material items are affected by being given an explanation showing feature ``parts" in the image. 
This may raise ethical concerns as feature highlighting could give users the impression that \enquote{incorrect} classification's seem less incorrect, but this concern is likely due to the fact that some images could plausibly be labelled as multiple different classes. 
To elaborate, Fig.~\ref{fig:expt4}(A) shows an example of these Right* items where a picture of a \enquote{Pot Pie} (decorated with lemons) is misclassified as \enquote{Lemon}. 
Here, the saliency explanation shows an image of a lemon and the \enquote{pulp of the lemon} as the feature that influenced the classification. 
When people see this explanation, it shows what the CNN is focusing on (i.e., the \enquote{Lemon} instead of the \enquote{Pot Pie}), leading them to rate the CNN classification as more correct (which makes sense), an effect which does not occur without using feature ``parts'' in the explanation.

\section{Conclusion} 
This work advances two novel case-based approaches to (i) highlight important regions in a test image and (ii) link these regions back to corresponding relevant regions in the training data. 
Unlike previous work, our approach is not constrained to specific architectures and can highlight multiple regions in an ANN agnostic fashion. 
Large scale ablation testing revels that the proposed approaches are more faithful to the model and select more relevant regions in the training data than previous solutions.
Results from a large scale user study indicate the utility of highlighting feature parts instead of just providing a relevant training example. 
Specifically, the study showed that the explanations help to appropriately calibrate people's understanding of how correct a classifier is on ambiguous test images.

\bibliographystyle{named}
\bibliography{ijcai23}

\end{document}